\pdfoutput=1

\documentclass[10pt,twocolumn,letterpaper]{article}

\usepackage[pagenumbers]{cvpr} 

\usepackage{graphicx}
\usepackage{amsmath}
\usepackage{amssymb}
\usepackage{booktabs}

\usepackage[pagebackref,breaklinks,colorlinks]{hyperref}

\usepackage[capitalize]{cleveref}
\crefname{section}{Sec.}{Secs.}
\Crefname{section}{Section}{Sections}
\Crefname{table}{Table}{Tables}
\crefname{table}{Tab.}{Tabs.}

\def\cvprPaperID{*****} 
\def\confName{CVPR}
\def\confYear{2023}

\begin{document}

\title{EmotiEffNet Facial Features in Uni-task Emotion Recognition in Video at ABAW-5 competition}

\author{Andrey V. Savchenko\textsuperscript{1,2}\\
\textsuperscript{1}Sber AI Lab\\
Moscow, Russia\\
\textsuperscript{2}HSE University\\
Laboratory of Algorithms and Technologies for Network Analysis, Nizhny Novgorod, Russia\\
{\tt\small avsavchenko@hse.ru}
}
\maketitle

\begin{abstract}
  In this article, the results of our team for the fifth Affective Behavior Analysis in-the-wild (ABAW) competition are presented. The usage of the pre-trained convolutional networks from the EmotiEffNet family for frame-level feature extraction is studied. In particular, we propose an ensemble of a multi-layered perceptron and the LightAutoML-based classifier. The post-processing by smoothing the results for sequential frames is implemented. Experimental results for the large-scale Aff-Wild2 database demonstrate that our model achieves a much greater macro-averaged F1-score for facial expression recognition and action unit detection and concordance correlation coefficients for valence/arousal estimation when compared to baseline.
\end{abstract}

\section{Introduction}
\label{sec:intro}
The affective behavior analysis in-the-wild (ABAW) problem is an essential part of many intelligent systems with human-computer interaction~\cite{kollias2019face,kollias2021distribution}. It can be used in online learning to recognize student satisfaction and engagement~\cite{savchenko2022classifying}, understand users' reactions to advertisements, analyze online event participants' emotions, video surveillance~\cite{sokolova2018organizing}, etc. Despite significant progress of deep learning in image understanding, video-based prediction of human emotions is still a challenging task due to the absence of large emotional datasets without dirty/uncertain labels. 

To speed up progress in this area, a sequence of ABAW workshops and challenges has been launched~\cite{kollias2020analysing,kollias2021analysing,kollias2022abaw4}. They introduced several tasks of human emotion understanding based on large-scale AffWild~\cite{kollias2019deep,zafeiriou2017aff} and AffWild2~\cite{kollias2019expression,kollias2021affect} datasets. The recent ABAW-5 competition~\cite{kollias2023abaw} contains an extended version of the Aff-Wild2 database for three uni-task challenges, namely, (1) prediction of two continuous affect dimensions, namely, valence and arousal (VA); (2) facial expression recognition (FER); and (3) detection of action units (AU), i.e., atomic facial muscle actions. It is strictly required to refine the model by using only annotations for a given task, i.e., the multi-task learning on the VA, FER, and AU labels of the AffWild2 dataset is not allowed. As emotions can rapidly change over time, frame-level predictions are required.

The above-mentioned tasks have been studied in the third ABAW challenge~\cite{kollias2022abaw} held in conjunction with CVPR 2022, hence, there exist several promising solutions for its participants. The baseline for VA prediction is a ResNet-50 pre-trained on ImageNet with a (linear) output layer that gives final estimates for
valence and arousal~\cite{kollias2023abaw}. Much better results on validation and test sets were achieved by EfficientNet-B0~\cite{savchenko2022cvprw} pre-trained on AffectNet~\cite{mollahosseini2017affectnet} from HSEmotion library~\cite{savchenko2022hsemotion}. An ensemble approach with the Gated Recurrent Unit (GRU) and Transformer~\cite{karpov2022exploring} combined using Regular Networks (RegNet)~\cite{Nguyen_2022_CVPR} took the third place for this task. The runner-up was a cross-modal co-attention model for continuous emotion recognition using visual-audio-linguistic information based on
ResNet-50 for spatial encoding and a temporal convolutional network (TCN) for temporal encoding~\cite{ZhangSu_2022_CVPR}. Finally, the winning solution utilized two types of encoders to capture the temporal context information in the video (Transformer and LSTM)~\cite{Meng_2022_CVPR}.

 The baseline for the FER task is a VGG16 network with fixed convolutional weights, pre-trained on the VGGFACE dataset~\cite{kollias2023abaw}. The second place was taken by an ensemble of multi-head cross-attention networks (Distract your Attention Network, DAN)~\cite{Jeong_2022_CVPR}. 
 A unified transformer-based multimodal framework
for AU detection and FER that uses InceptionResNet visual features and DLN-based audio features~\cite{Zhang_2022_CVPR} took first place in this competition.

Similarly to FER, the baseline for AU detection is a VGGFACE network~\cite{kollias2023abaw}. A visual spatial-temporal transformer-based model and a convolution-based audio model to extract action unit-specific features were proposed in~\cite{Wang_2022_CVPR}. An above-mentioned ensemble approach of GRU and Transformer with RegNets~\cite{Nguyen_2022_CVPR} took third place. Slightly better results were achieved by the IResnet100 network that utilized feature pyramid networks and single-stage headless~\cite{Jiang_2022_CVPR}. The winner is again the InceptionResNet-based audiovisual ensemble of the Netease Fuxi Virtual Human team~\cite{Zhang_2022_CVPR}

In this paper, we propose a novel pipeline suitable for all three tasks of ABAW in the video. The unified representation of a facial emotion state is extracted by a pre-trained lightweight EmotiEffNet model~\cite{savchenko2021emotions}. These convolutional neural networks (CNN) are tuned on external AffectNet dataset~\cite{mollahosseini2017affectnet}, so the facial embeddings extracted by this neural network do not learn any features that are specific to the Aff-Wild2 dataset~\cite{kollias2019expression,kollias2021affect}. Several blending ensembles are studied based on combining embeddings and logits at the output of these models for each video frame~\cite{savchenko2022cvprw,savchenko2023hse}. In addition to MLP (multi-layer perceptron), we examine classifiers from the LightAutoML (LAMA) framework~\cite{vakhrushev2021lightautoml}.

The remaining part of the paper is organized as follows. The proposed workflow is presented in Section~\ref{sec:2}. Its experimental study for three tasks from the fifth ABAW challenge is provided in Section~\ref{sec:3}. Finally, Section~\ref{sec:4} contains the conclusion and discussion of future studies.
 
\section{Proposed approach}\label{sec:2}

\begin{figure}[t]
 \centering
 \includegraphics[width=0.95\linewidth]{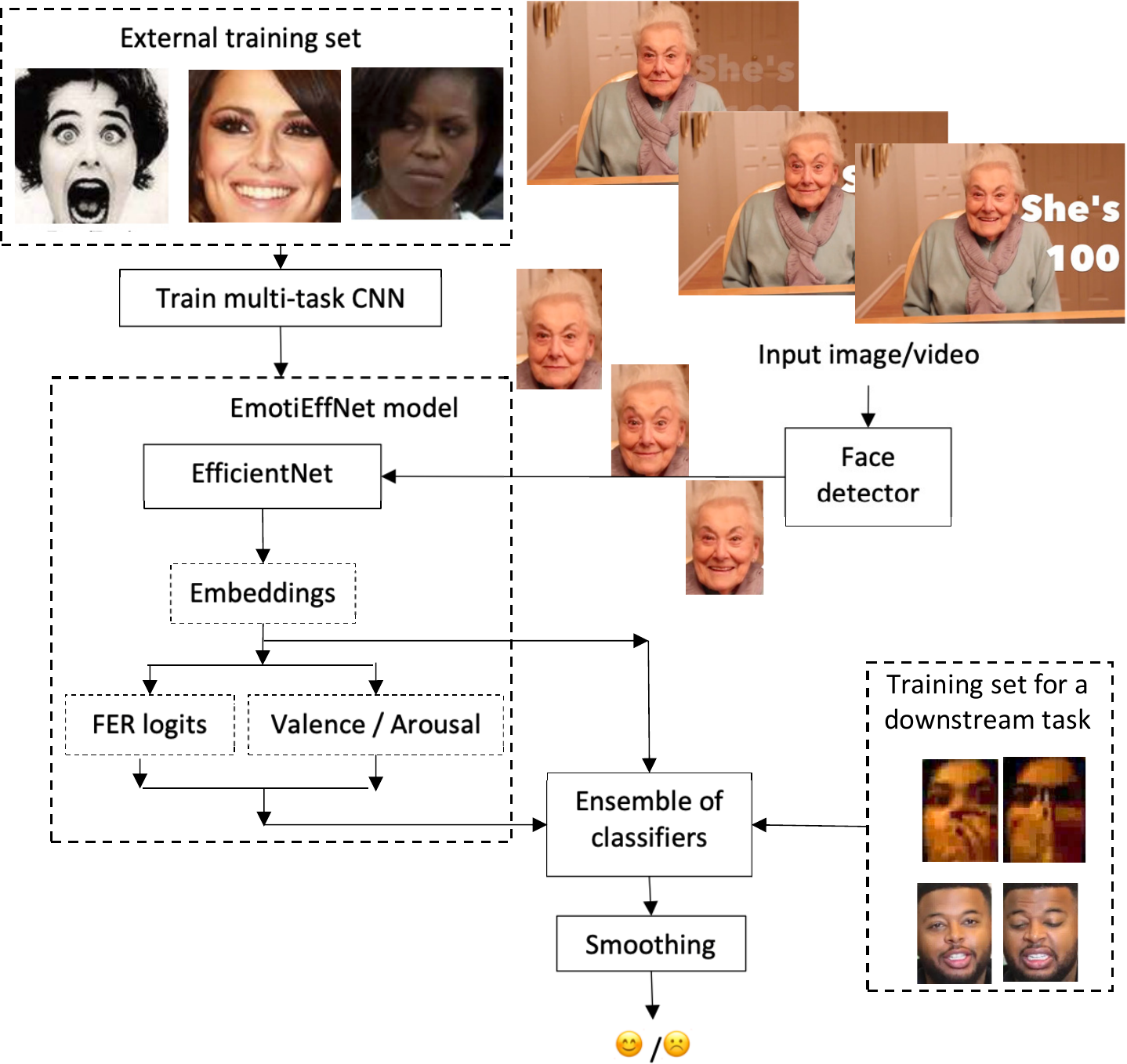}
 \caption{Proposed workflow for the video-based facial emotion analysis.}
 \label{fig:1}
\end{figure}

In this Section, the novel workflow for emotion recognition in video is introduced (Fig.~\ref{fig:1}). At first, the faces are detected with an arbitrary technique, and the representations of affective behavior are extracted from each face by using EfficientNet CNN from HSEmotion library~\cite{savchenko2022hsemotion}, such as EmotiEffNet-B0~\cite{savchenko2022cvprw} or the winner of one of the tasks from ABAW-4, namely, MT-EmotiEffNet-B0~\cite{savchenko2023hse}. These models were trained for face identification on the VGGFace2 dataset. Next, they were finetuned to recognize facial expression and, in case of multi-task MT-EmotiEffNetmodel, predict valence/arousal from a static photo by using the AffectNet dataset~\cite{mollahosseini2017affectnet}.

For simplicity, let us assume that every $t$-th frame of the video contains a single facial image $X(t)$, where $t \in \{1,2,..., T\}$ and $T$ is the total number of frames~\cite{savchenko2015statistical}. These images are resized and fed into the EmotiEffNet PyTorch models to obtain $D$-dimensional embeddings $\mathbf{x}(t)$, eight-dimensional logits for 8 facial expressions $\mathbf{l}(t)$ from AffectNet (Anger, Contempt, Disgust, Fear, Happiness, Neutral, Sadness, Surprise) and valence $V(t)\in [-1;1]$ (how positive/negative a person is) and arousal $A(t)\in [-1;1]$ (how active/passive a person is)~\cite{kollias2023abaw}.

Next, these facial representations are used to solve an arbitrary downstream task. In this paper, we examine three problems from the ABAW-5 competition, namely (1) VA prediction (multi-output regression); (2) FER (multi-class classification); and (3) AU detection (multi-class multi-label classification).

The supervised learning case is assumed where a training set of $N>1$ pairs $(X_n,y_n), n=1,2,...N$ is available. Here a facial image $X_n$ from the video frame and associated with corresponding labels $y_n$. Here are the details about each task:
\begin{enumerate}
  \item VA estimation data contains a training set with 356 videos and 1653757 frames, and a validation set with 76 videos and 376323 frames
  \item Training and validation sets for FER contain 248 videos (585317 frames) and 70 videos (280532 frames), respectively. Each frame is associated with one of eight imbalanced classes (Neutral, Anger, Disgust, Fear, Happiness, Sadness, Surprise, or Other).
  \item AU detection challenge consists of a training set (295 videos, 1356694 frames) and a validation set (105 videos, 445836 frames) with 12 highly-imbalanced labels: AU 1 (inner brow raiser, AU 2 (outer brow raiser), AU 4 (brow lowerer), AU 6 (cheek raiser), AU 7 (lid tightener), AU 10 (upper lip raiser), AU 12 (lip corner puller), AU 15 (lip corner depressor), AU 23 (lip tightener), AU 24 (lip pressor), AU 25 (lips part), and AU 26 (jaw drop)~\cite{kollias2023abaw}.
\end{enumerate}

At first, every training example $X_n$ is fed into the same CNN to obtain embeddings $\mathbf{x}_n$, FER logits $\mathbf{l}_n$ and valence/arousal $V_n, A_n$. In this paper, the following classifiers are trained: MLP and ensemble models trained via the LAMA library~\cite{vakhrushev2021lightautoml}. The latter tries to find the best pre-processing, classifiers and their ensembles, and post-processing for an arbitrary classification or regression task. Due to computational complexity and poor metrics obtained after 10 minutes of AutoML search, we do not process embeddings $\mathbf{x}(t)$ here. The input of LAMA is a concatenation of logits $\mathbf{l}(t)$, valence $V(t)$, and arousal $A(t)$ at the output of the last layer of EmotiEffNets. 

The MLP is trained with the TensorFlow 2 framework similarly to~\cite{savchenko2022cvprw}. VA is better predicted using only logits and valence/arousal by an MLP without a hidden layer and two outputs with $tanh$ activation functions trained to maximize the mean estimate of the Concordance Correlation Coefficient (CCC) for valence $CCC_V$ and arousal $CCC_A$. 

FER and AU detection are solved similarly by feeding embeddings or logits into the MLP with one hidden layer. In the former case, eight outputs with softmax activations were added, and the weighted sparse categorical cross-entropy is used to fit the classifier. In addition, we examined the possibility to finetune the whole EmotiEffNet CNN on the training set of this challenge using PyTorch source code from the HSEmotion library. 

The output layer for the AU detection task contains 12 units with sigmoid activation functions, and the weighted binary cross-entropy loss was optimized. The final prediction is made by matching the outputs with predefined thresholds. It is possible to either set a fixed threshold (0.5 for each unit) or choose the best threshold for each action unit to maximize F1-score on a validation set. The classifier predicts the class label that corresponds to the maximal output of the softmax layer.

In all tasks, it is possible to build a simple blending decision rule to combine several different classifiers (LightAutoML, MLP, finetuned model) and input features (embeddings or logits from pre-trained model). Moreover, pre-trained models were used to make predictions for VA and FER tasks. In the former case, the valence $V(t)$ and arousal $A(t)$ predicted by MT-EmotiEffNet~\cite{savchenko2023hse} were directly used to make a final decision (hereinafter ``pre-trained VA only"). In the second case due to the difference in classes, namely, absence of contempt emotion and the presence of the state ``Other" in the AffWild2 dataset, we preliminary apply the MLP classifier to make a binary decision (Other/non-Other). If the predicted class label is not equal to Other, predictions of the pre-trained model from other 7 basic facial expressions are used, i.e., the label that corresponds to the maximal logit (hereinafter ``pre-trained logits").

The final decision in the pipeline (Fig.~\ref{fig:1}) is made by smoothing predictions (class probabilities for classification tasks and predicted valence/arousal for regression problem)~\cite{savchenko2015statistical} for individual frames by using the box filter with kernel size $2k+1$. Here $k$ is a hyperparameter chosen to maximize performance metrics on the validation set. In fact, we compute the average predictions for the current frame, previous $k$ frames, and next $k$ frames~\cite{savchenko2022cvprw}.

\section{Experimental results}\label{sec:3}

Let us discuss the results of our workflow (Fig.~\ref{fig:1}) for three tasks from the fifth ABAW challenge~\cite{kollias2023abaw}. The training source code to reproduce the experiments for the presented approach is publicly available\footnote{\url{https://github.com/HSE-asavchenko/face-emotion-recognition/blob/main/src/ABAW/}}. 

\subsection{Valence-Arousal Prediction}

\begin{figure}[t]
 \centering
 \includegraphics[width=0.95\linewidth]{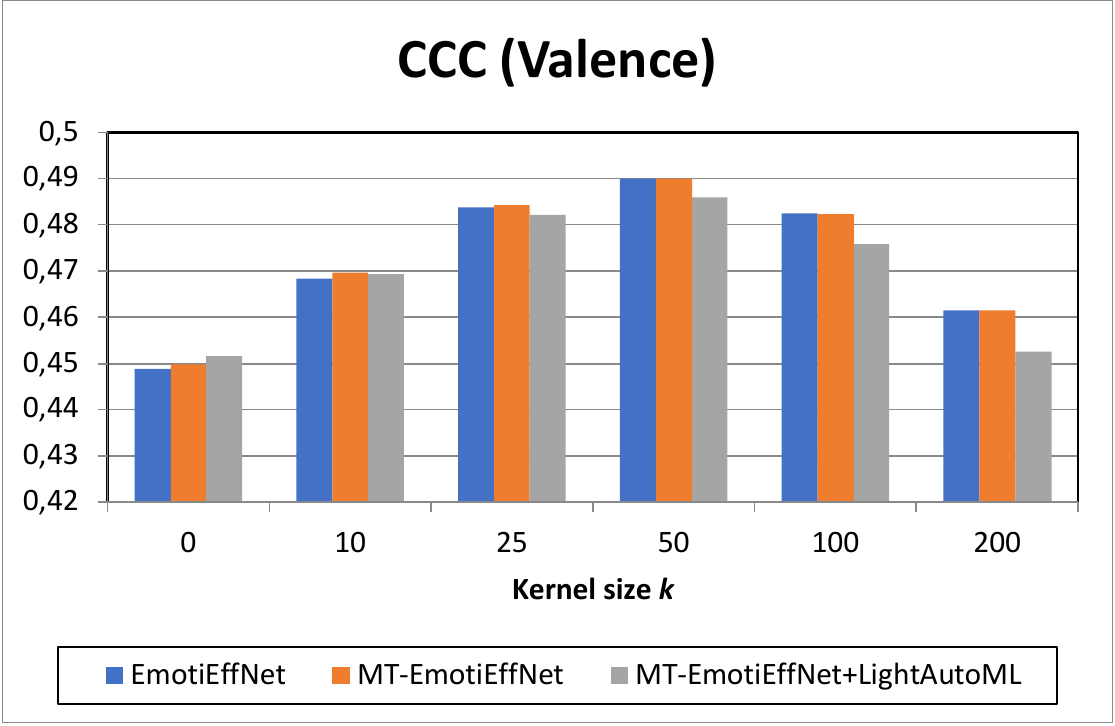}
 \caption{Dependence of CCC for valence prediction on the smoothing kernel size $k$.}
 \label{fig:ccc_v}
\end{figure}
\begin{figure}[t]
 \centering
 \includegraphics[width=0.95\linewidth]{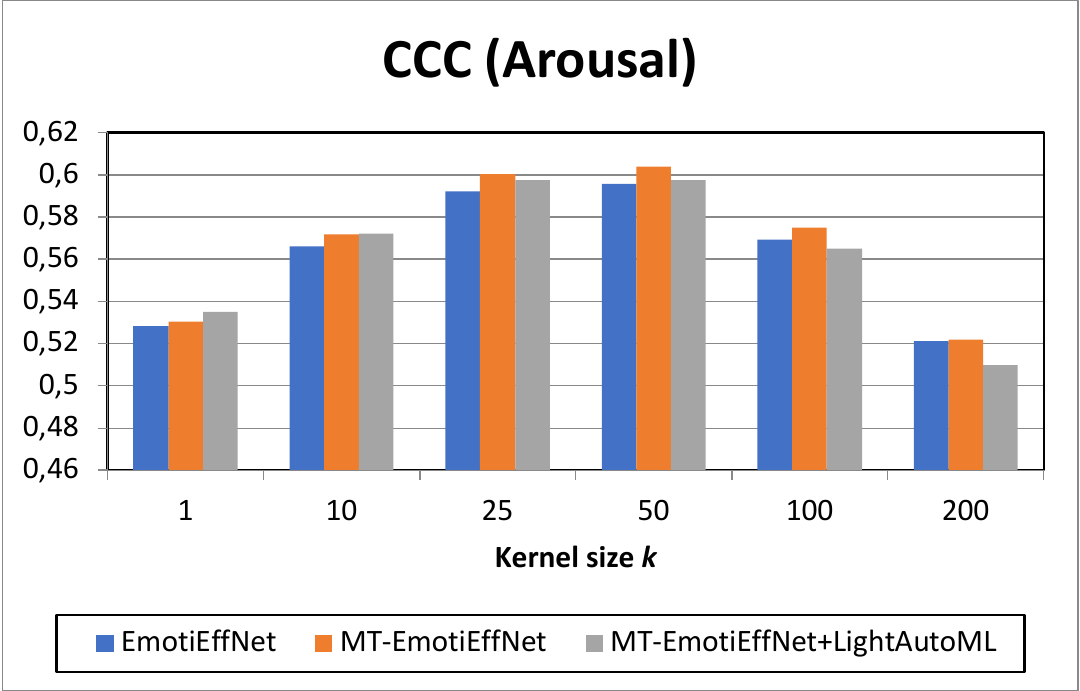}
 \caption{Dependence of CCC for arousal prediction on the smoothing kernel size $k$.}
 \label{fig:ccc_a}
\end{figure}

A comparison of our workflow based on EmotiEffNet-B0 and MT-EmotiEffNet-B0~\cite{savchenko2023hse} with previous results on the Valence-Arousal estimation challenge is presented in Table~\ref{tab:va}. We use official performance metrics from the organizers: CCC for valence, arousal, and their average value $P_{VA}=(CCC_V+CCC_A)/2$. 

\begin{table*}
 \centering
 \begin{tabular}{ccccc}
 \toprule
 Model & Method & CCC\_V & CCC\_A& Mean CCC $P_{VA}$ \\
 \midrule
 ResNet-50 & Baseline~\cite{kollias2023abaw} & 0.31 & 0.17 & 0.24\\
 EfficientNet-B0 &HSE-NN~\cite{savchenko2022cvprw} & 0.449 & 0.535 & 0.492\\
 GRU + Attention & PRL ensemble~\cite{Nguyen_2022_CVPR} & 0.437 & 0.576 & 0.507 \\
 Resnet-50/TCN & FlyingPigs audio/video ensemble~\cite{ZhangSu_2022_CVPR} & 0.450 & 0.651 & 0.551\\
 Transformer & Situ-RUCAIM3 audio/video ensemble~\cite{Meng_2022_CVPR} & 0.588 & 0.669 & 0.627 \\
 \hline
 & Pre-trained logits, LightAutoML & 0.373 & 0.433 & 0.403 \\
 	& Pre-trained logits, MLP & 0.444 & 0.521 & 0.483 \\
	MT-EmotiEffNet& Pre-trained VA only & 0.404 & 0.248 & 0.326 \\
 	& Pre-trained logits, MLP + LightAutoML & 0.447 & 0.526 & 0.487 \\
 & Pre-trained logits, MLP, smoothing & 0.490 & 0.604 & 0.547 \\
 	& Pre-trained logits, MLP + LightAutoML, smoothing & 0.486 & 0.597 & 0.542 \\
 \hline
 & Pre-trained logits, LightAutoML & 0.369 & 0.431 & 0.400 \\
 EmotiEffNet & Pre-trained logits, MLP & 0.443 & 0.519 & 0.482 \\
 & Pre-trained logits, MLP, smoothing & 0.490 & 0.596 & 0.543 \\
 \hline
 EmotiEffNet + & Pre-trained logits, MLP & 0.450 & 0.530 & 0.490 \\
 MT-EmotiEffNet &Pre-trained logits, MLP, smoothing & 0.494 & 0.607 & 0.550 \\
 \bottomrule
 \end{tabular}
 \caption{Valence-Arousal Challenge Results on the Aff-Wild2’s validation set.}
 \label{tab:va}
\end{table*}

\ifdefined\DEBUG
lama (0.37270991068427667, 0.43333181223266476, 0.4030208614584707)
EmotiEffNet
logreg: (0.4436768680313978, 0.5195607326809414, 0.48161880035616955)
new: (0.4433857992438458, 0.5198270222465867, 0.48160641074521626)

MT-EmotiEffNet
logreg:(0.44436749382135277, 0.5207544767285259, 0.48256098527493935)
only va predictions (0.40439760158295573, 0.24760572372463172, 0.3260016626537937)

ensemble scores: (0.44347719917685247, 0.5154618245315408, 0.4794695118541966)
ensemble lama:0.8
(0.4468021962997973, 0.5264224337187418, 0.4866123150092695)

smooth
vgaf:
1 (0.44875087823993354, 0.5283116909318734, 0.48853128458590345)
10 (0.4684252179690231, 0.5658810666512365, 0.5171531423101298)
25 (0.4838215785829268, 0.5921912567094488, 0.5380064176461878)
50 (0.48999702278941953, 0.5955375478706236, 0.5427672853300216)
100 (0.4824520565285041, 0.5694326567832584, 0.5259423566558812)
200 (0.46156615573225407, 0.5209989880709679, 0.49128257190161095)

mtl:
1 (0.4497322853156013, 0.5302893781650251, 0.4900108317403132)
10 (0.46963786100578786, 0.5716260813822904, 0.5206319711940391)
25 (0.4843495213418185, 0.6004171233716765, 0.5423833223567476)
50 (0.48997545654787067, 0.6038755689432846, 0.5469255127455777)
100 (0.48241128098915237, 0.5749223688356685, 0.5286668249124105)
200 (0.46145788788100156, 0.5220036569168273, 0.4917307723989144)

mtl+lama
1 (0.451674645802567, 0.5350759616300828, 0.4933753037163249)
10 (0.4693974742540891, 0.5721432029699063, 0.5207703386119977)
25 (0.482241973924982, 0.5974393614123437, 0.5398406676686629)
50 (0.4858742077572039, 0.5973650198783172, 0.5416196138177606)
100 (0.4759241499975304, 0.5649415306338962, 0.5204328403157132)
200 (0.452538972675504, 0.50987857208748, 0.481208772381492)

vgaf+mtl
1 (0.45467808539350846, 0.5384962410483103, 0.49658716322090934)
10 (0.4735034168577817, 0.5771482697977821, 0.5253258433277819)
25 (0.4880783142242917, 0.6040955053246918, 0.5460869097744918)
50 (0.4935198754791566, 0.6068185446719148, 0.5501692100755357)
100 (0.4853385612499721, 0.5783366485114904, 0.5318376048807312)
200 (0.4638047085529785, 0.5265323923684997, 0.4951685504607391)
\fi

Here, we significantly improved the results of the baseline ResNet-50~\cite{kollias2023abaw}: our CCC is higher up to 0.19 and 0.53 for valence and arousal, respectively. Moreover, our best ensemble model is characterized by 0.05 greater CCC\_V and 0.07 greater CCC\_A when compared to the best previous usage of EmotiEffNet models~\cite{savchenko2022cvprw}. The LightAutoML classifier is worse than simple MLP, but their blending achieves one of the top results. 

One of the most valuable hyperparameters in our pipeline is the kernel size $k$ of the median filter. The dependence of validation CCCs on $k$ for valence and arousal is shown in Fig.~\ref{fig:ccc_v} and Fig.~\ref{fig:ccc_v}, respectively. As one can notice, the highest performance is reached by rather high values of $k$ (25...50), i.e., 51...101 predictions should be averaged for each frame.

\subsection{Facial Expression Recognition}
\begin{figure}[t]
 \centering
 \includegraphics[width=0.95\linewidth]{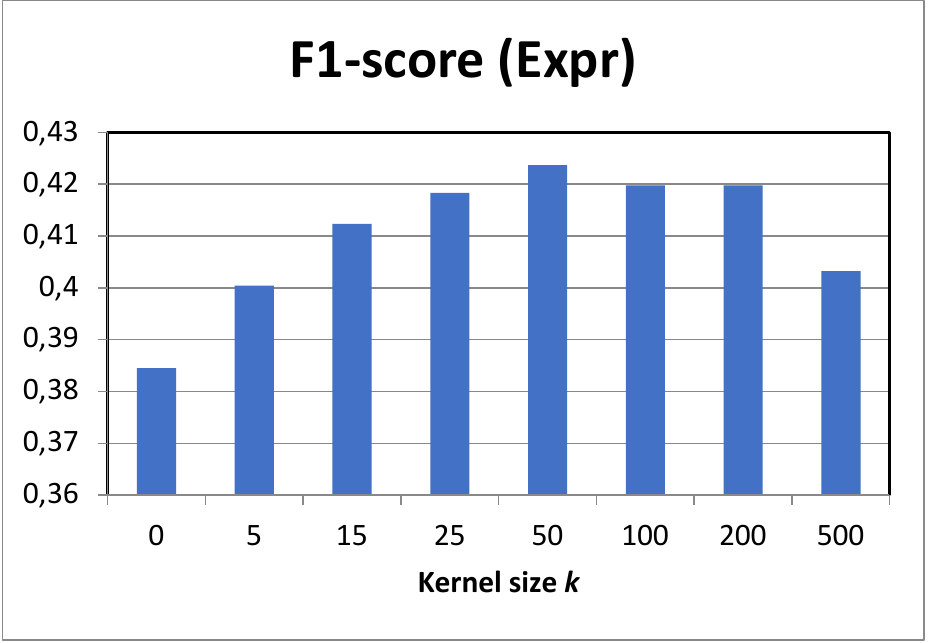}
 \caption{Dependence of F1-score for FER on the smoothing kernel size $k$.}
 \label{fig:expr}
\end{figure}

The macro-averaged F1-scores $P_{EXPR}$ and classification accuracy estimated on the official validation set for various FER techniques are shown in Table~\ref{tab:expr}. The value of $P_{EXPR}$ depending on the kernel size $k$ for our several best classifiers is presented in Fig.~\ref{fig:expr}. 

\begin{table*}
 \centering
 \begin{tabular}{cccc}
 \toprule
 DNN & Method details & F1-score $P_{EXPR}$ & Accuracy \\
 \midrule
 VGGFACE & Baseline~\cite{kollias2023abaw} & 0.23 & -\\
\hline
EfficientNet-B0 &HSE-NN~\cite{savchenko2022cvprw} & 0.402 & - \\
DAN (ResNet50) & IXLAB ensemble~\cite{Jeong_2022_CVPR} & 0.346 & - \\
 InceptionResNet & Netease Fuxi Virtual Human ensemble~\cite{Zhang_2022_CVPR} & 0.394 & - \\
\hline
	MT-EmotiEffNet& Pre-trained embeddings for aligned faces & 0.293 & 0.403 \\
 &Pre-trained embeddings for cropped faces & 0.336 & 0.447 \\
 \hline
& Pre-trained embeddings for aligned faces & 0.304 & 0.474 \\
 &Pre-trained embeddings for cropped faces & 0.384 & 0.495 \\
EmotiEffNet& Pre-trained embeddings with smoothing & 0.432 & 0.546 \\
& Pre-trained logits & 0.327 & 0.426 \\
 & Finetuned model for cropped faces & 0.380 & 0.484 \\
 \hline
& Pre-trained embeddings + logits, frame-level & 0.396 & 0.502 \\
EmotiEffNet& Pre-trained embeddings + logits, smoothing & 0.431 & 0.546 \\
 ensemble& Pre-trained + finetuned, frame-level & 0.405 & 0.524 \\
& Pre-trained + finetuned, smoothing & 0.433 & 0.557 \\
 \bottomrule
 \end{tabular}
 \caption{Expression Challenge Results on the Aff-Wild2’s validation set.}
 \label{tab:expr}
\end{table*}

In addition to cropped faces provided by the organizers of this challenge, we used here small (112x112) cropped\_aligned photos. As one can notice, the quality of FER on the former is much better, so we do not use aligned faces in other experiments. The proposed approach makes it possible to increase F1-score on 20\% and 3\% when compared to the baseline VGGFACE~\cite{kollias2023abaw} and the previous usage of EfficientNet models~\cite{savchenko2022cvprw}. Again, a large kernel size (50...200) of the mean filter is required to provide the best-possible F1-score (Fig.~\ref{fig:expr})

Surprisingly, MT-EmotiEffNet~\cite{savchenko2023hse} is up to 5\% worse than EmotiEffNet, though the former was significantly more accurate on the multi-task learning challenge from ABAW-4 competition~\cite{kollias2022abaw4}. It is important to emphasize that the F1-score of our best ensemble (43.3\%) is approximately equal to the best single model (43.2\%), though the difference in accuracy is significant (55.7\% vs 54.6\%). Nevertheless, we achieved the greatest validation F1-score, which is 4\% higher than the F1-score of the ABAW-3 winner team (Netease Fuxi Virtual Human)~\cite{Zhang_2022_CVPR}.

\ifdefined\DEBUG
EmotiEffNet
finetuning: Acc 0.4836667474655298 0.3801234273402112
ensemble (weight=0.8): 0.8 (0.4051360355253487, 0.5243287753268789)

smoothing (delta=50)
EmotiEffNet: Acc: 0.5462585373504627 F1: 0.431761062689951
Ensemble (weight=0.8): Acc: 0.5574194744271598 F1: 0.4326929159405485

lama Acc: 0.21001525672650537 F1: 0.1019832060515383

EmotiEffNet-B0 scores:
OTHER=True (280532,) (280532,) 3698 (0.2574340560684253, 0.32562417121754383, [0.47261988187005705, 0.08159349444926674, 0.16895322458679918, 0.08013937282229966, 0.462785267165087, 0.4738317407716553, 0.28230688149375094, 0.037242585388486384])

OTHER=False (174088,) (174088,) 1883 (0.38315520007157966, 0.5219142043104636, [0.5941131445271728, 0.11673050704408232, 0.2563962686160057, 0.1146315414738341, 0.6221359518916038, 0.574566439059762, 0.40351254788859714])

scores only: 0.3270045426963257, 0.42616885061240783
ensemble: 0.5(0.3962179837417037, 0.5023704960574907)
smooth: Acc: 0.546454593415368 F1: 0.4312567829587687

EmotiEffNet
aligned
112 Acc: 0.36997808749986594 F1: 0.21352704404593095
224 Acc: 0.47395701146384794 F1: 0.3045351324629094
cropped Acc: 0.49544436998274705 F1: 0.3844313672414089

MT-EmotiEffNet
aligned 112 Acc: 0.36310406828978836 F1: 0.21416588541497295
cropped Acc: 0.44668344431294826 F1: 0.3355545672821115

smooth
0 Acc: 0.49544436998274705 F1: 0.3844313672414089
5 Acc: 0.5147255928022472 F1: 0.40041647458913604
15 Acc: 0.5290234269174283 F1: 0.41233375265784433
25 Acc: 0.5352187985684342 F1: 0.41830667436931435
50 Acc: 0.542572683330244 F1: 0.4236698679610459
100 Acc: 0.5408224373689989 F1: 0.4197549134058517
200 Acc: 0.5397387820284317 F1: 0.41970605922798776
500 Acc: 0.5307736728786734 F1: 0.4031952312654022
\fi

\subsection{Action Unit Detection}

\begin{table*}
 \centering
 \begin{tabular}{ccc}
 \toprule
 DNN & Method details & F1-score $P_{AU}$ \\
 \midrule
 VGGFACE & Baseline~\cite{kollias2023abaw} & 0.39\\
 EfficientNet-B0 &HSE-NN~\cite{savchenko2022cvprw} & 0.548\\
 Transformer & STAR-2022 audio/video ensemble~\cite{Wang_2022_CVPR} & 0.523 \\
 GRU + Attention & PRL ensemble~\cite{Nguyen_2022_CVPR} & 0.544 \\
 IResNet & SituTech~\cite{Jiang_2022_CVPR} & 0.735 \\
 InceptionResNet & Netease Fuxi Virtual Human ensemble~\cite{Zhang_2022_CVPR} & 0.525 \\
\hline
& LightAutoML, threshold 0.5 & 0.389 \\
 & LightAutoML, best thresholds & 0.472 \\
& MLP, threshold 0.5 & 0.494 \\
 MT-EmotiEffNet & MLP, best thresholds & 0.525 \\
 & MLP, smoothing & 0.533 \\
 & MLP + LightAutoML, threshold 0.5 & 0.529 \\
 & MLP + LightAutoML, best thresholds & 0.533 \\
 & MLP + LightAutoML, smoothing & 0.543 \\
\hline
& LightAutoML, threshold 0.5 & 0.397 \\
 & LightAutoML, best thresholds & 0.477 \\
& MLP, threshold 0.5 & 0.508 \\
 EmotiEffNet & MLP, best thresholds & 0.537 \\
& MLP, smoothing & 0.545 \\
 & MLP + LightAutoML, threshold 0.5 & 0.502 \\
 & MLP + LightAutoML, best thresholds & 0.542 \\
 & MLP + LightAutoML, smoothing & 0.554 \\
 \hline
 & Ensemble, threshold 0.5 & 0.520 \\
 EmotiEffNet + MT-EmotiEffNet & Ensemble, best thresholds & 0.544 \\
& Ensemble, smoothing & 0.553 \\
 \bottomrule
 \end{tabular}
 \caption{Action Unit Challenge Results on the Aff-Wild2’s validation set.}
 \label{tab:au}
\end{table*}

\ifdefined\DEBUG
EmotiEffNet
features
dense: 0.507568027779308
(0.5367145284583944, 0.65
 MT-EmotiEffNet
 0.49428877376095776
(0.524806956530541, 0.7000000000000001

lama
EmotiEffNet
lama 0.3967976517739536
0.47712935942351486
ensemble w=0.6
0.5019551098097399
(0.536135316544336, 0.65
or 0.5416010326422609

MT-EmotiEffNet
0.39161635065008743
0.47801455675819343, 0.25000000000000006

ensemble w=0.6
0.5289615773736197
(0.5328972153003431, 0.525

EmotiEffNet+MT-EmotiEffNet
0.6000000000000001
0.5198255232950532
(0.543975201831992, 0.6249999999999999, array([0.57638351, 0.46688322, 0.57987711, 0.6332615 , 0.7558231 ,
    0.755736 , 0.73109244, 0.36942135, 0.20410174, 0.23078234,
    0.84263162, 0.38170849]), array([0.7, 0.8, 0.7, 0.5, 0.5, 0.5, 0.6, 0.8, 0.7, 0.8, 0.3, 0.6]), array([0.88472443, 0.91312501, 0.87501458, 0.80686172, 0.7981141 ,
    0.81579774, 0.86405764, 0.95661409, 0.94202801, 0.94938722,
    0.78110337, 0.84139011]))

smoothing:
EmotiEffNet
0 0.5367145284583944 
3 0.5456568248392965 
5 0.5453739302190261 
7 0.5445119128957571 
10 0.5416391816305005
15 0.5349712162447015
20 0.5294193258656613
25 0.5234582906500872
50 0.5008177570051234 

ensemble with lama
0 0.5416010326422609 
3 0.5528112024399152 
5 0.5539261586365561 
7 0.5535949604016204 
10 0.5520381951293492
15 0.5485896969154594
20 0.544495477255726 
25 0.5397615003471172
50 0.5235874148818092

 MT-EmotiEffNet
0 0.524806956530541 
3 0.5325899361581461
5 0.532621354151279 
7 0.5309452680793919 
10 0.5269493143058434 
15 0.5187949627232132 
20 0.5112837798767936 
25 0.5059059359752085 
50 0.48470887742529567

ensemble with lama
0 0.5289615773736197 
3 0.5403141268526156 
5 0.5423988784220879 
7 0.5427907663334933 
10 0.5418239193970038
15 0.5375846932762096
20 0.5339041037335406
25 0.5311038538198062
50 0.5181744018193561 

best
0.5: 7 0.5427202485732338
thresholds: 5 0.5432475964435367 

EmotiEffNet+MT-EmotiEffNet
0 0.543975201831992 
3 0.5527115882582081 
5 0.5528469599376815 
7 0.5514931236426248 
10 0.5490788247920431
15 0.5433385627648376
20 0.5380824835642478
25 0.5329111662587908
50 0.5109522680195706

best
0.5: 15 0.535933703990885 
thresholds 5 0.5528469599376815 
\fi

\begin{figure}[t]
 \centering
 \includegraphics[width=0.95\linewidth]{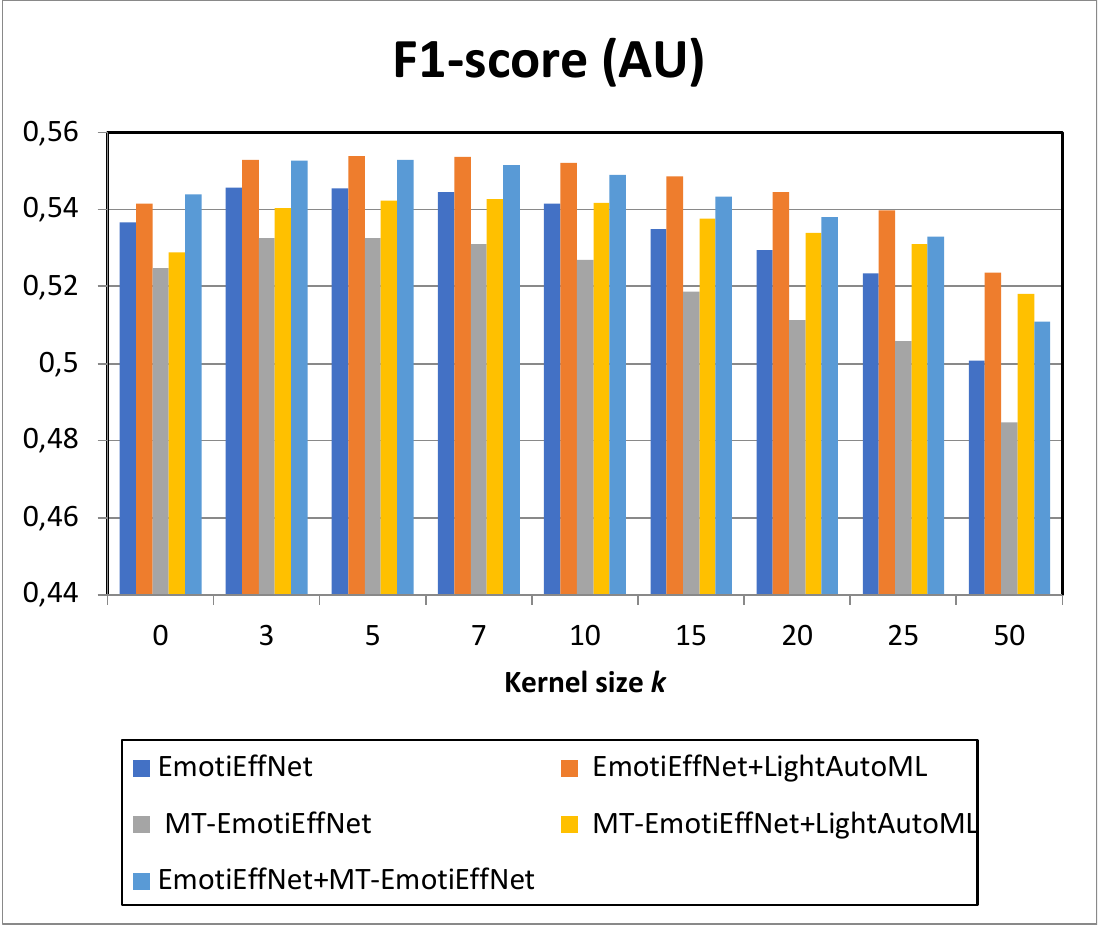}
 \caption{Dependence of F1-score for AU detection on the smoothing kernel size $k$.}
 \label{fig:au}
\end{figure}

In the last Subsection, macro-averaged F1-score $P_{AU}$ is estimated for the multi-label classification of action units. The estimates of performance metrics on the validation set are shown in Table~\ref{tab:au} and Fig.~\ref{fig:au}. In contrast to previous experiments, the kernel size $k$ should be much lower (3...5) to achieve the best performance. Indeed, at least one action unit is rapidly changed in typical scenarios. 

The LightAutoML ensemble is again slightly worse than a simple MLP, but their blending leads to excellent results. Our best model is 16\% better than the baseline, though we increase the F1-score of EmotiEffNet compared to its previous usage~\cite{savchenko2022cvprw} on 1\%. However, only the second-place winner team (SituTech) has a higher F1 score on the validation set. Finally, the choice of thresholds can definitely improve the AU detection quality, though it is possible that the current choice of a threshold for each action unit using the validation set is not optimal.

\section{Conclusion and future works}\label{sec:4}

We proposed the video-based emotion recognition pipeline (Fig.~\ref{fig:1}) suitable for a wide range of affective behavior analysis downstream tasks. It exploits the pre-trained EmotiEffNet models to extract representative emotional features from each facial frame. Experiments on datasets from the fifth ABAW challenge showed the benefits of our workflow when compared to the baseline CNNs~\cite{kollias2023abaw} and previous application of EfficientNet models~\cite{savchenko2022cvprw}. For example, our best models achieved average CCC for VA estimation $P_{VA}=55\%$ and macro-averaged F1 scores for FER $P_{EXPR}=43.3\%$ and AU detection $P_{AU}=55.3\%$.

Our workflow (Fig.~\ref{fig:1}) uses only facial modality and frame-level features. Hence, it is possible to significantly improve the quality metrics by combining it with audio processing~\cite{savchenko2013phonetic,Zhang_2022_CVPR} and/or temporal models~\cite{Meng_2022_CVPR,ZhangSu_2022_CVPR}. Another direction for future research is an increase in decision-making speed by using sequential inference~\cite{savchenko2021fast} and processing of video frames.

\textbf{Acknowledgements}. The work is supported by RSF (Russian Science Foundation) grant 20-71-10010.

{\small
\bibliographystyle{ieee_fullname}
\bibliography{paper_abaw5}
}

\end{document}